\documentclass[a4paper,fleqn]{cas-dc}

\usepackage[numbers]{natbib}
\usepackage{upgreek}
\usepackage{bbding}
\usepackage{hhline}
\usepackage{balance}
\usepackage{picins}
\usepackage{diagbox}

\usepackage{tabu}                     
\usepackage{multirow}                 
\usepackage{multicol}                 
\usepackage{multirow}                
\usepackage{float}                    
\usepackage{makecell}                 
\usepackage{booktabs}                 
\usepackage{color}
\usepackage{multicol}
\usepackage[justification=centering]{caption}

\def\tsc#1{\csdef{#1}{\textsc{\lowercase{#1}}\xspace}}
\tsc{WGM}
\tsc{QE}
\tsc{EP}
\tsc{PMS}
\tsc{BEC}
\tsc{DE}

\begin{document}
\let\WriteBookmarks\relax
\def\floatpagepagefraction{1}
\def\textpagefraction{.001}
\shortauthors{CV Radhakrishnan et~al.}

\title [mode = title]{ADD: An Automatic Desensitization Fisheye Dataset for Autonomous Driving}                      

\tnotetext[1]{Corresponding author.}

\author[1]{Zizhang Wu}[type=editor,
                        auid=000,
                        bioid=1]
\cormark[1]


 \address[1]{the Zongmu Technology (Shanghai) Co., Ltd, Building 10, Zhangjiang Artificial Intelligence Island, Lane 55, Chuanhe Road, Pudong New Area, Shanghai, China}

\author[1]{Xinyuan Chen}

\author[1,2]{Hongyang Wei}[%
   ]


 \address[2]{the Xinjiang University, College of Software, No. 14 Shengli Road, Urumqi, Xinjiang Uygur Autonomous Region, China}

\author%
[1]
{Fan Song}



\author[1]{Tianhao Xu}
\cormark[1]

\nonumnote{E-mail address: zizhang.wu@zongmutech.com (Z. Wu),  lan.chen@zongmutech.com (X. Chen), weihy@stu.xju.edu.cn (H. Wei), fan.song@zongmutech.com (F. Song), tobias.xu@zongmutech.com (T. Xu).
 }

\begin{abstract}
Autonomous driving systems require many images for analyzing the surrounding environment.
%
However, there is fewer data protection for private information among these captured images, such as pedestrian faces or vehicle license plates, which has become a significant issue.  
%
In this paper, in response to the call for data security laws and regulations and based on the advantages of large Field of View(FoV) of the fisheye camera, we build the first \textbf{A}utopilot \textbf{D}esensitization \textbf{D}ataset, called \textbf{ADD}, and formulate the first deep-learning-based image desensitization framework, to promote the study of image desensitization in autonomous driving scenarios.
The compiled dataset consists of 650K images, including different face and vehicle license plate information captured by the surround-view fisheye camera.
It covers various autonomous driving scenarios, including diverse facial characteristics and license plate colors.
%
Then, we propose an efficient multitask desensitization network called \textbf{DesCenterNet} as a benchmark on the \textbf{ADD} dataset, which can perform face and vehicle license plate detection and desensitization tasks. 
Based on \textbf{ADD}, we further provide an evaluation criterion for desensitization performance, and
%
%
%
extensive comparison experiments have verified the effectiveness and superiority of our method on image desensitization.
\end{abstract}




\maketitle
\section{Introduction}\label{sec1}
With the massive explosion of big data, the in-depth improvement of algorithms, and the continuous development of hardware technology, artificial intelligence technology represented by deep learning has ushered in a new round of prosperity. Although the current stage is still in the era of quasi-artificial intelligence, and artificial intelligence cannot be granted legal subject qualification, a series of problems caused by artificial intelligence technology still need to be responded to and regulated by law.

The technology and application of artificial intelligence have greatly enhanced the value of data. However, the problems of illegal acquisition, illegal trading, and illegal leakage of data are also particularly prominent and have seriously affected the data security of every member of society. According to the regulation proposed by \cite{bib1036} and the viewpoint of Ling \cite{bib1040}, sensitive data generated by intelligent driving vehicles should be masked before updating to other devices. Data use must first undergo desensitization processing, that is, deprivacy processing of data to protect sensitive information, which can not only effectively use data but also ensure the security of data use. Therefore, data desensitization is necessary to filter sensitive information before further analysis. 

The rapid development of autonomous driving is accompanied by the generation of many images from the surrounding environment.
These images will be served for subsequent tasks, such as pedestrian detection\cite{bib92,bib93,bib94,bib95,bib96,bib2}, route planning\cite{bib97,bib98,bib99,bib100}, and automatic parking\cite{bib101,bib102,bib1,bib3,bib4,bib5}.
%
To achieve this, most of these images will be directly transferred into remote servers or computing clouds for further analysis, typically driven by deep-learning-based image processing and analysis techniques.
%
However, these images frequently include sensitive personal information, such as faces or license plates. This private information is typically used for data transmission, storage, and analysis in current autonomous driving systems, with no privacy protection.
%

\begin{figure}
\centering
\includegraphics[width=0.45\textwidth]{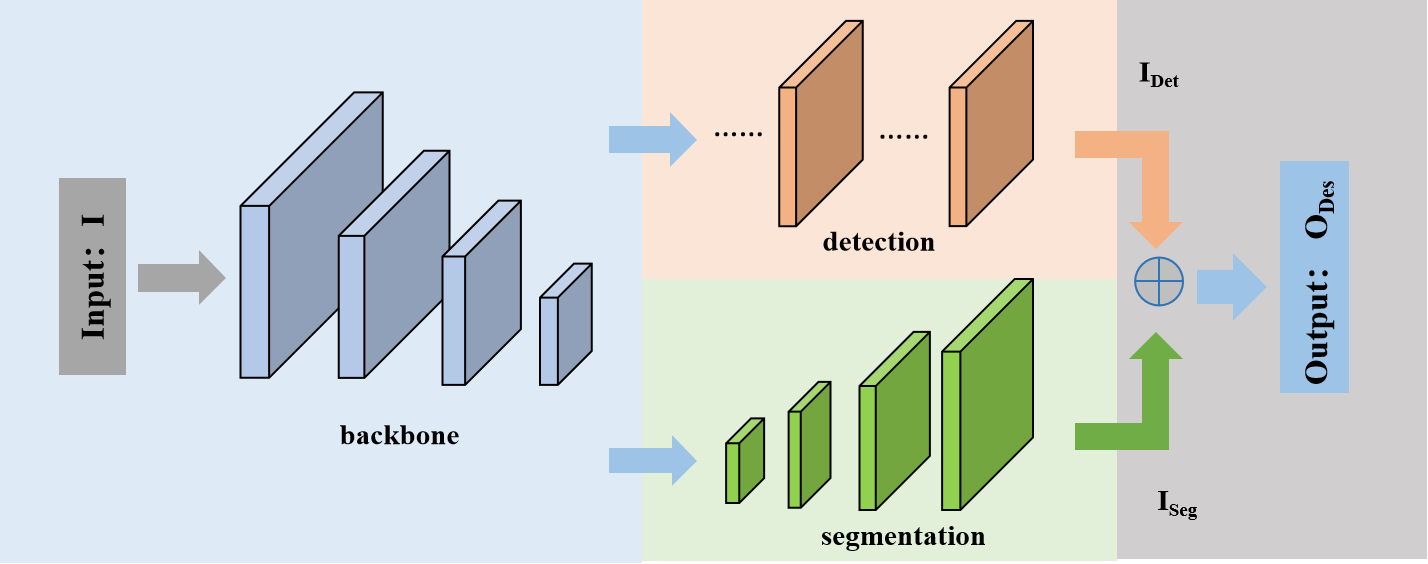}
\caption{Desensitization task sketch.}
\label{fig:firstimage}    
\end{figure}

In autonomous driving, pedestrian and vehicle information are vital and fundamental, so ensuring the privacy of pedestrian faces and license plate numbers is especially critical.
%
In this paper, we propose a task for desensitization of face and vehicle license plate information in automatic driving. 
Data desensitization~\cite{bib8,bib9} detects sensitive information and then masks or obscures it before data transmission or analysis.
Given the original or background image $I$, a desensitized image $O_{Des}$ can be modelled as a combination of a detection module $I_{Det}$ with Joint desensitization method $M$ and segmentation module $I_{Seg}$,
\begin{equation}
O_{des}=M(I_{seg},I_{det})  \quad if\quad Iou_{(I_{seg},I_{det})}>0.5
\end{equation}
The $M$ is a Joint desensitization method where each module $I$ deals with a desensitized region if $ IOU_{I_{seg},I_{det}}>0.5$, and otherwise belongs to the negative sample. $IOU_{I_{seg},I_{det}}$ calculates iou with $I_{det}$ by finding the minimum bounding box of the segmented region.
$O_{Des}$ can be any image to cover the private information, such as a cartoon icon or grey rectangle square, as shown in Fig.~\ref{fig:firstimage}.



 
%

Current datasets and benchmarks for face and vehicle license plate detection, which frequently include urban, field, or highway scenes,
include Wider Face\cite{bib80},CCPD\cite{bib53},CRPD\cite{bib54}, etc. 
However, existing research lacks a dataset of face and vehicle license plate desensitization in autonomous driving scenarios and does not propose the relevant task of face and vehicle license plate information desensitization. 
%
 %
%
%
To complement this field, we propose the first \textbf{A}utopilot \textbf{D}esensitization \textbf{D}ataset (\textbf{ADD}) and a desensitization framework to promote the study of the desensitization of face and vehicle license plate information. Table \ref{compare:dataset}  summarizes our benchmarks and provides a comparison to existing datasets. 


%

First, we compile a new large-scale fisheye dataset that is provided to facilitate studies dealing with face and vehicle license plate detection and desensitization.
In addition, it is subdivided into three sub-datasets: \textbf{F}ace \textbf{D}esensitization dataset(\textbf{FD}), \textbf{V}ehicle license plate \textbf{D}esensitization dataset(\textbf{VD}), and \textbf{mix}ed \textbf{D}esensitization dataset(\textbf{mixD}), each serving different training or evaluation purposes. The face desensitization dataset considers face information of different ages, genders, hairstyles, expressions, and accessories. The vehicle license plate desensitization dataset considers vehicle license plate information of different colors. The hybrid desensitization dataset contains both face and vehicle license plate information, which is more informative. To ensure the diversity of the data, we collected data from different scenes, periods, and attributes. Unlike other face and vehicle license plate recognition datasets\cite{bib53,bib54,bib56,bib57,bib58,bib59,bib60,bib62,bib63,bib64,bib65,bib66}, ours focuses more on the desensitization of face and vehicle license plate information and provides 650K images of autonomous driving scenarios.

We propose a new practical two-stage multitask desensitization network (MDN) for face and vehicle license plate desensitization tasks to better apply the dataset to real autonomous driving scenarios. 
MDN can detect the face and vehicle license plate and the desensitization task of the face and vehicle license plate information. We also established a new desensitization evaluation index to verify the desensitization effect. We divide the facial features into three regions and assign different weights to represent the importance of facial features (IOFF), similar to AP \cite{bib79} and other indicators, to evaluate the desensitization effect.

Our contributions are summarized as follows:
\begin{itemize}
\item  To the best of our knowledge, the first fisheye dataset for face and vehicle license plate desensitization contains 650K images with adequate face information and vehicle license plate information in the autonomous driving scenario.\\
\item  We build the face and vehicle license plate desensitization benchmark on the \textbf{ADD} dataset and propose a novel MDN to achieve detection and desensitization tasks.
\end{itemize}

The rest of the paper is organized as follows. Sec. 2 introduces the relevant methods, while Sec. 3 covers the \textbf{ADD} dataset in detail. Then we provide an overview of the proposed desensitization network in Sec. 4, and Sec. 5 describes the comparative analysis. Finally, Sec. 6 concludes this paper.

\begin{table*}
\centering
\caption{Comparison of current State-of-the-Art Benchmarks and Datasets.}
\label{compare:dataset}

\resizebox{\linewidth}{!}{
\begin{tabular}{cccccccc}
\multicolumn{8}{l}{\textbf{detection and segmentation}}  \\ \hline
Datasets    & Setting        & Sensor type    & Ground Label                      & \#categories & avg. \#labels/category & \#images & Resolution \\ \hline
COCO\cite{bib79}        & indoor/outdoor & camera         & pedestrian/car                    & 80           & 18K                    & 220K     & 800*600    \\
VOC\cite{bib24}         & indoor/outdoor & camera         & pedestrian/car                    & 20           & 22K                    & 44K      & 500*375    \\
KITTI\cite{bib25}       & outdoor        & camera         & pedestrian/car                    & 2            & 80K                    & 160K     & 1242*375   \\ \hline
\textbf{ADD}         & indoor/outdoor & fisheye camera & pedestrian/car/vehicle license plate/face & 4            & 162K                   & 650K     & 1920*1280  \\ \hline
\multicolumn{8}{l}{}
\end{tabular}}
\begin{tabular}{ccccccc}
\multicolumn{7}{l}{\textbf{vehicle license plate detection and recognition}}  \\ \hline
Datasets    & Setting        & Sensor type    & Ground Label                      & \#categories  & \#images & Resolution \\ \hline
CCPD\cite{bib53}        & indoor/outdoor & camera         & vehicle license plate                     & 1                                  & 250K     & 720*1160   \\
CPRD\cite{bib54}        & indoor/outdoor & camera         & vehicle license plate                     & 1                                  & 30K      & 640*640    \\
ReID\cite{bib56}        & outdoor        & camera         & vehicle license plate                     & 1                                  & 76K      & 200*40     \\ \hline
\multicolumn{7}{l}{}
\end{tabular}

\begin{tabular}{cccccc}
\multicolumn{6}{l}{\textbf{face detection and recognition}}  \\ \hline
Datasets    & Setting        & Sensor type    & Ground Label                      & \#categories  & \#images  \\ \hline
WebFace260M\cite{bib57} & real           & camera         & face                              & 1                                  & 260M               \\
MS-Celeb-1M\cite{bib61} & real           & camera         & face                              & 1                                  & 10M               \\
MF2\cite{bib62}         & real           & camera         & face                              & 1                                 & 4.7M              \\ 
LFW\cite{bib65}        & real           & camera         & face                              & 1                                 & 13K               \\ \hline
\multicolumn{6}{l}{}
\end{tabular}
\resizebox{\linewidth}{!}{
\begin{tabular}{cccccc} 
\multicolumn{6}{l}{\textbf{autonomous driving datasets}}  \\ \hline
Datasets    & Setting        & image type    & Ground Label                    & numFisheyeCameras  & Resolution \\ \hline
WoodScape\cite{bib1029}        & real & fisheye image         & pedestrian/car                    & 4                & 1280*966    \\
SynWoodScape\cite{bib1030}         & synthetic & fisheye image         & pedestrian/car                    & 4           & 1280*966    \\
KITTI 360\cite{bib1031}      & real        & fisheye image         & pedestrian/car                    & 2                 & 1400*1400   \\ 
FisheyeCityScapes\cite{bib1032}         & real & fisheye image         & pedestrian/car                    & 1                 & 600*600    \\
THEODORE\cite{bib1033}         & synthetic & fisheye image         & pedestrian/car                    & 1                 & 1024*1024    \\
OmniScape\cite{bib1034}         & synthetic & fisheye image         & pedestrian/car                    & 2                 & 1024*1024    \\ \hline
\textbf{ADD}         & real & fisheye image         & pedestrian/car/vehicle license plate/face                    & 4                 & 1920*1280    \\ \hline
\end{tabular}}
\end{table*}

\section{Related Work}
\subsection{Image processing-based datasets}
Various datasets have been compiled to meet the needs of deep learning models for data. In terms of image processing, there are object detection and segmentation datasets \cite{bib79,bib24,bib25}, image denoising datasets \cite{bib46,bib47,bib48}, image deraining datasets \cite{bib49,bib50,bib51}, image definition datasets \cite{bib52}, etc. The popularity of fisheye cameras has also produced many automatic driving fisheye datasets\cite{bib1029,bib1030,bib1031,bib1032,bib1033,bib1034}.
%
These datasets, combined with deep learning models, have greatly improved image processing performance. However, there is no dataset for image desensitization. Therefore, we propose new vehicle license plates and face desensitization datasets for image desensitization. 
It is very different from common vehicle license plate recognition datasets \cite{bib53,bib54,bib56} and face recognition datasets \cite{bib57,bib58,bib59,bib60,bib62,bib63,bib64,bib65,bib66}. It is not simply detection and identification. Our desensitization dataset aims to combine the desensitization method with the convolutional neural network to achieve the desensitization of the face and vehicle license plate.

\subsection{Detection and segmentation-based tasks}
With the rapid development of object detection \cite{bib13,bib14,dong2022lightweight,fu2022learning} and semantic segmentation technology \cite{bib83,bib85,bib86,bib87}, research on face detection \cite{bib15,bib16,7553523}, car plate detection \cite{bib17,bib18}, and instance segmentation \cite{bib19,bib20,bib1037,bib1038} has increased, and increasingly more research on pedestrian detection  \cite{bib21,bib22} and vehicle detection \cite{bib23} based on depth learning has effectively solved the density problems and occlusion in the natural environment. Both faces and license plates are vital signs of pedestrians and vehicles and have unique attributes representing their parents, which have a unique role in identifying specific pedestrians and vehicles. Therefore, it is necessary to use pedestrian and vehicle detection and segmentation methods to enhance the extraction and recognition of face and license plate information. The desensitization task is similar to instance segmentation tasks, such as Mask R-CNN \cite{bib1037} and FCIS \cite{bib1038}. The difference is that it proposes a new desensitization evaluation method for the critical features of the face and license plate, and the effect is particularly prominent in the instance segmentation task.

\subsection{Traditional desensitization-based tasks}
The explosive growth of data promotes the protection of sensitive information in various forms of data. The traditional recognition method of sensitive information based on rules and regular expressions \cite{bib6,bib7} requires much expert knowledge, has poor mobility, and the recognition pattern is relatively rigid. Unstructured data (text, image, etc.) desensitization technology based on deep learning and machine learning came into being \cite{bib8,bib9}. However, tasks applied to image desensitization are very few \cite{bib11,bib12}. For an image to be desensitized, mage areas having content related to the content of a partial image area that needs to be desensitized are automatically recognized based on the partial image area. Alternatively, an image area to be desensitized is automatically recognized based on a desensitization rule corresponding to a service scenario with which the image to be desensitized is associated. Alternatively, a complete set of image areas to be desensitized is acquired by automatic extension based on a selected partial image area. This paper combines object detection with semantic segmentation and realizes the desensitization task of face and license plate information. Compared with the previous desensitization methods, the combined desensitization method proposed in this paper is more novel and significant.

\section{Autopilot Desensitization Dataset}
In this section, we introduce our \textbf{ADD} dataset in detail, including the data collection and preprocessing, annotation protocols, informative statistics, dataset characteristics, and dataset application.
\subsection{Data Collection and Preprocessing}
%
As shown in Table \ref{tab:Dataset Characteristic} (a), we collect a total of three cities, three scenarios (parking, street, and highway), and two periods (morning and evening) and capture 200 videos. These videos last from 1 hour to 6 hours with an average of 2 hours, where the density of faces or vehicle license plates is at least 3 (The density of the dataset is shown in Table \ref{tab:Dataset Characteristic} (a)).
%
We perform frame extraction processing on the video (with a frame rate of 2 FPS after frame extraction), store the video as frame images, and select images containing facial and license plate instances. For high-quality images, we restrict the visible range of the fisheye camera and further remove distorted and blurred face and vehicle license plate images. It is worth noting that we do not distort the images collected by the fisheye camera and filter the images we need through human eye judgment. We also enhance the image data, such as copy-paste\cite{bib1035} that can increase the density of faces/license plates in an image. Taking facial copy-paste as an example, we select a basic background image that may contain 0 pedestrian targets or at least one pedestrian target. By pasting some diverse features of pedestrians (as shown in Table \ref{tab:Dataset Characteristic} (b)) into the background image, we enrich its pedestrian and facial diversity. The same applies to the license plate copy-paste protocol. Then, the collected data will be divided according to the density of people and vehicles to ensure the diversity of \textbf{ADD}. Finally, for the convenience of other studies, we divide the data into face, license plate sub-datasets and a mixed sub-dataset with face and license plate.

\subsection{Data Annotation}
\begin{figure*}
\centering
\includegraphics[width=1.0\textwidth]{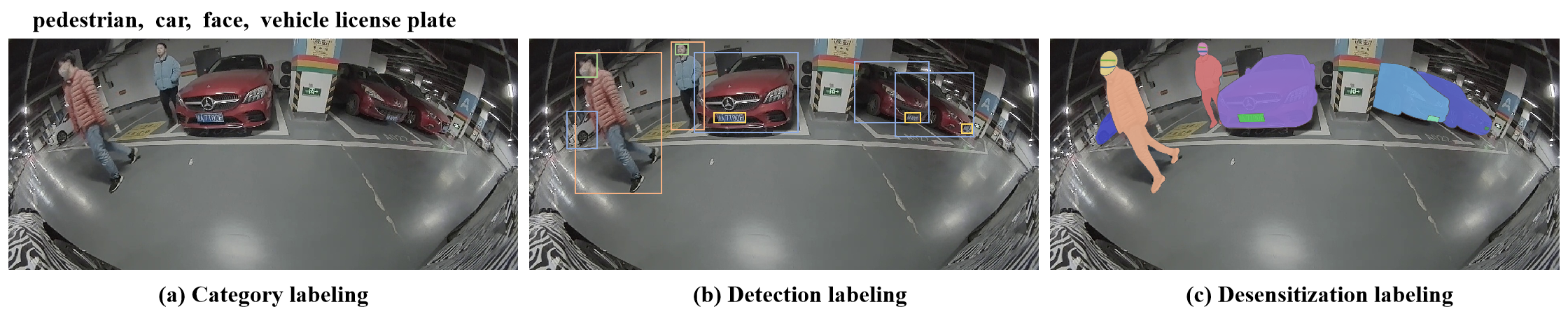}
\caption{Our image annotation pipeline is split into three primary worker tasks: (a) Labelling the categories present in the image, (b) Locating and marking all instances of the labelled categories, and (c) Segmenting each object instance.}
\label{fig2}
\end{figure*}

We next describe how we annotated our image collection. Due to our desire to label over 250K images, the design of a cost-efficient yet high-quality annotation pipeline was critical, as Fig.\ref{fig2} shows. Our marking task is distributed to a total of 5 marking engineers in the form of crowdsourcing, and each engineer is responsible for marking a batch of data. The marking personnel are all professional and use the marking tools to mark effectively and, finally, return to the marking engineer for acceptance. General acceptance is to check whether our requirements for labelled pipes are met.
The labelling method of this paper is similar to the COCO dataset\cite{bib79} detection and segmentation labelling method. We performed data cleaning in the collection phase, reducing the time and cost of labelling personnel. The first task in annotating our dataset is determining which object categories are present in each image, Fig. \ref{fig2}(a). In the next stage, we label the location and category of the object, as shown in Fig. \ref{fig2}(b). Finally, we add masks to each target, but in particular, we divide the face mask into three areas: the area above the eyes, the area from the eyes to the nose, and the area below the nose. The regions help us effectively evaluate face desensitization, Fig. \ref{fig2}(c). The basic principle of all labelling is that only visible targets can be labelled, but if the visible target area is less than half of the target itself, it will not be marked. Additionally, we do not cover all pedestrians' continuous moving processes for redundant annotations but select some highly visible targets as tracking objects and generate pedestrian trajectory annotation information and video sequence information (the annotated temporal sequences length of 40). Finally, we collect the first large-scale fisheye desensitization dataset for autopilot application.


\begin{table*}
\caption{Properties and statistics of \textbf{ADD}}
\label{tab:Dataset Characteristic}
\centering
\resizebox{\linewidth}{!}{
\begin{tabular}{c|c|c|ccc|c}

\multicolumn{7}{l}{\textbf{(a) Statistics}}  \\ \hline
\multirow{2}{*}{Dataset}                     & \multirow{2}{*}{Sub-dataset}               & \multirow{2}{*}{Image resolution} & \multicolumn{3}{c|}{\#images} & \multirow{2}{*}{density(/image)} \\ \cline{4-6}
                                             &                                            &                                   & train     & val     & test    &                                  \\ \hline
\multirow{3}{*}{Autopilot Desensitization} & Face Desensitization                     & \multirow{3}{*}{1920×1280}        & 100k     & 60k   & 40k   & 5$\sim$6                         \\
                                             & Vehicle license plate Desensitization               &                                   & 100k     & 60k   & 40k   & 3$\sim$4                         \\
                                             & Face and vehicle license plate Desensitization(mixD) &                                   & 125k    & 75k   & 50k   & 8$\sim$10                        \\ \hline
\multicolumn{7}{l}{}
\end{tabular}}

\resizebox{\linewidth}{!}{
\begin{tabular}{c|c|c|c|c}
\multicolumn{5}{l}{\textbf{(b) Properties}}  \\ \hline
Scenario                 & Time period              & Category              & Attribute   & Explanation                                             \\ \hline
\multirow{2}{*}{Parking} & \multirow{3}{*}{Daytime} & \multirow{5}{*}{Face} & Age         & Faces of different ages                                 \\
                         &                          &                       & Gender      & Faces of different genders                              \\ 
\multirow{2}{*}{Street}  &                          &                       & Hairstyle   & Faces with different hairstyles                         \\ 
                         & \multirow{3}{*}{Night}   &                       & Expression  & Faces with different expressions                        \\ 
\multirow{2}{*}{Highway} &                          &                       & Accessories & Such as hats, glasses, masks, headphones, scarves, etc. \\ \cline{3-5} 
                         &                          & Vehicle license plate            & Color       & vehicle license plates in different colors                          \\ \hline
\end{tabular}}

\end{table*}

\subsection{Dataset Description}
\begin{figure*}
\includegraphics[width=1.0\textwidth]{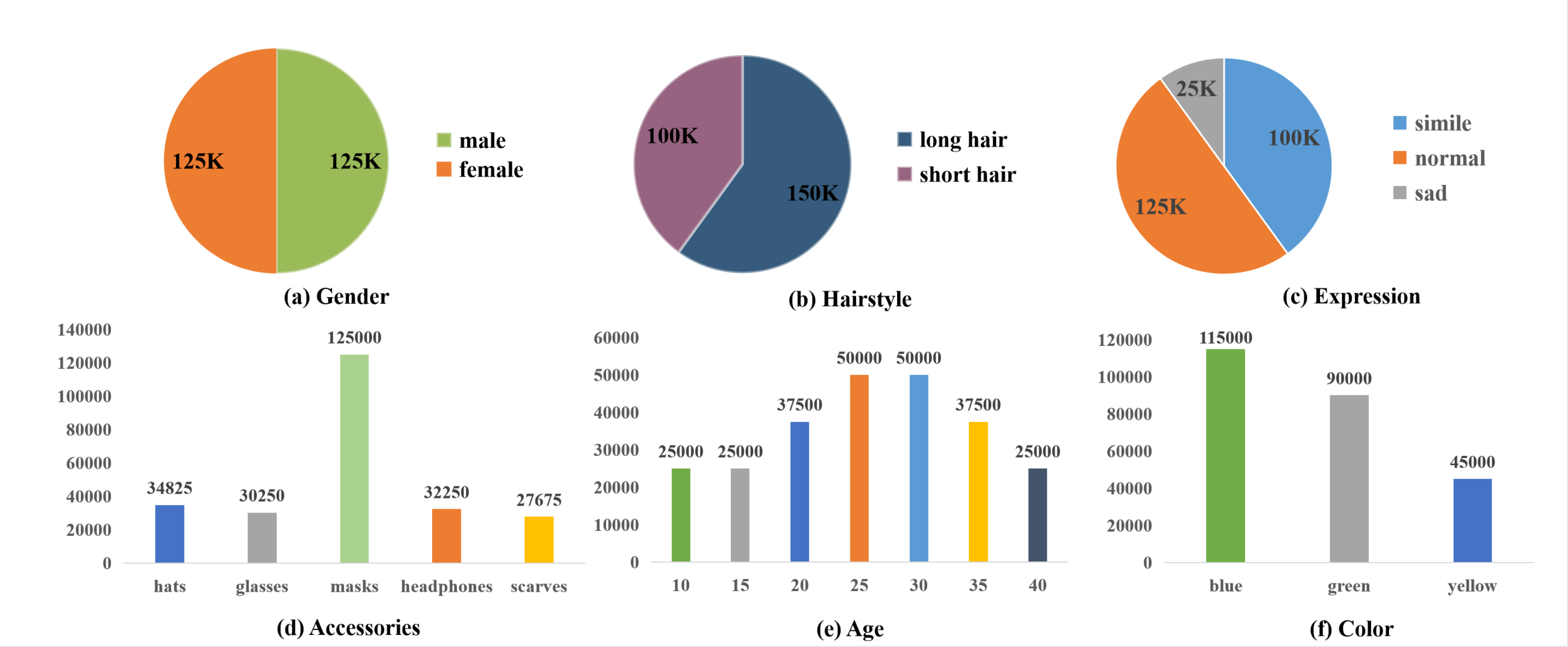}
\centering
\caption{Statistics of various attributes in the \textbf{mixD} subdataset. Attributes of Faces include (a) Gender. (b) Hairstyle. (c) Facial expressions. (d) Different accessories. (e) Age. (f) Different colors are the main attributes of the vehicle license plate.}
\label{fig:data_attribute}    
\end{figure*}

To allow different sub-datasets to be used for the training of different desensitization tasks, we specifically divide them into three sub-datasets to meet this premise: (1) \textbf{F}ace
\textbf{D}esensitization (\textbf{FD}); (2) \textbf{V}ehicle license plate \textbf{D}esensitization (\textbf{VD}); and (3) \textbf{mix}ed \textbf{D}esensitization (\textbf{mixD}), as shown in 
Table \ref{tab:Dataset Characteristic}(b).
A total of more than 650K images comprise three sub-datasets: \textbf{FD}, \textbf{VD}, and \textbf{mixD}, with amounts of 200K, 200K, and 250K, respectively. 
The image resolution is 1920x1280, 
and the average object density of each image is 5-6, 3-4, and 8-10 respectively. To complete the face and license plate desensitization task, we decided to use the \textbf{mixD} dataset for further desensitization research. To further describe the basic properties of different targets in the \textbf{mixD} dataset, we have made detailed statistics in Fig. \ref{fig:data_attribute}, and all the data meet the requirements of the actual scene as much as possible.


\subsection{Dataset Characteristics}
Our \textbf{ADD} dataset exhibits differences from existing datasets in image representation style, scenarios, quantity, and diversity. Below, we elaborate on the four main characteristics of our \textbf{ADD} dataset.
\\

\noindent{\textbf{Fisheye image representation.}}
The \textbf{ADD} dataset consists of fisheye images, different from the common pinhole images of public datasets. Fisheye images provide a larger field-of-view (FoV), which is more suitable for close-range and low-lying face and vehicle license plate detection and recognition.\\

\noindent{\textbf{Specifically for autopilot scenarios.}}
The \textbf{ADD} dataset focuses on the face and vehicle license plate desensitization of high-speed or low-speed parking scenarios in the autonomous driving field, which is different from the natural scene of public datasets. The environmental conditions in autonomous driving scenarios, such as dark and opaque conditions, greatly increase the difficulty of desensitization of face and vehicle license plates. Regarding research on the application of desensitization in autonomous driving scenarios, the \textbf{ADD} dataset can promote the study of face desensitization and vehicle license plate desensitization in the real world.\\

\noindent{\textbf{Great quantity.}}
Our \textbf{ADD} dataset obtains more than 650k data from more than 200-hour autopilot scene video clips. We constantly collect various scenarios, including facial and vehicle license plates, and finally reach more than hundreds of thousands of face and vehicle license plate data.\\

\noindent{\textbf{High quality and diversity.}}
Our \textbf{ADD} dataset covers three cities, three scenarios from different periods, and different face and vehicle license plate cases. In addition, we selected many people's facial attributes, including age and gender. We also selected many vehicle license plate attributes to include different colors. Finally, we carefully select high-quality, high-resolution images to ensure the dataset's advantages.\\

\subsection{Evaluation Metrics}
\subsubsection{Detection Evaluation}
In this paper, we first use AP, AP50, and AP75 to evaluate the detection performance of the model.
AP (Average Precision) is an essential measure of the accuracy of the object detection algorithm, 
by evaluating the object's position in the image and predicting the object class.

\begin{equation}
R \mathrm{eca} l l=\frac{T P}{\mathrm{T P}+\mathrm{F N}}
,
\text { Precision }=\frac{T P}{\mathrm{T P}+\mathrm{F P}}
\end{equation}

\begin{equation}
A P=\frac{\sum P_{c}}{N_{c_{-} \text {total }}}, c\in [0,C)
\end{equation}
where TP represents \textbf{T}rue \textbf{P}ositives (samples correctly classified as positive), FN is \textbf{F}alse \textbf{N}egatives (samples incorrectly classified as negative), and FP is false positives (samples incorrectly classified as positive).
$C$ is the number of categories in the dataset and
$P_{c}$ indicates the precision of category $c$, and $N_{c\_total}$ represents the total number of categories in the test set of category $c$. AP is obtained through Precision and Recall. AP50 is the average precision obtained when the detector threshold is more significant than 50\%. AP75 is the average precision obtained when the detector threshold exceeds 75\%. 

\subsubsection{Desensitization Evaluation}
 To achieve the goal of this task, we use evaluation metrics similar to the Intersection over Union (IoU). The difference is that IoU evaluates every region of the objects with the same weight, and we assign different weights to different regions according to the importance of the facial features and plate letters.

\begin{table}[h]
\caption{Three regional weights of average importance of facial features (IoFF)}
\label{tab:IoFF}
\centering

\begin{tabular}{c|c|c|c} 
\hline
region&above the eyes & from eyes to nose & below the nose \\
\hline
weight&25\% & 50\% & 25\% \\
\hline
\end{tabular}
\end{table}

For the face desensitization task, to wipe the critical facial feature against face recovery technologies, we evaluate the importance of different facial features by collecting 1000 faces and dividing them into three different regions: above the eyes, from eyes to nose, and below the nose. We mask each of them with three masking regions and test the similarity between the original face and the masked face using the network proposed by Schroff\cite{2015FaceNet}, and the confidence of the similarity is treated as the importance of the facial features, with a confidence ratio close to 1:2:1. Therefore, we assigned weights of the same ratio (as shown in Table 3).

Table \ref{tab:IoFF} shows the three regional weights of the Importance of Facial Features ($IoFF$). It is worth noting that the $IoFF$ positively correlates to the distance to the eyes, and the most critical region to identify a person is the eyes, which takes up to 50\% importance to identify a person. To calculate the final score of face desensitization, we sum all the scores of different masked regions to perform the same calculation of mean Average Precision (mAP) as illustrated in equation 5\cite{bib82}.

\begin{equation}
    IoFF=\left\{
    \begin{aligned}
    &\sum_{i=1}^{N} W_{f_i} * F_{IoU_i}& if\quad c = face \\
    &IoU          & if\quad c = plate \\
    \end{aligned}
    \right.
\end{equation}

\begin{equation}
mIOFF= \left\{
\begin{aligned}
&\frac{\sum_{i=1}^{N}IoFF\left ( i \right )}{N} & if\quad c = face \\
& IoFF  & if\quad c = plate\\
    \end{aligned}
    \right.
\end{equation}

where W$_{f_i}$ and F$_{IoU}$ are the i$_{th}$ weight and IOU of the corresponding facial features, respectively, c is the category of the prediction result, and N is the number of face regions. 

For the vehicle license plate desensitization task, since the importance of each letter is the same, we assign the same weight to each of them and perform the same calculation as mAP.

\begin{figure*}
\includegraphics[width=1.0\textwidth]{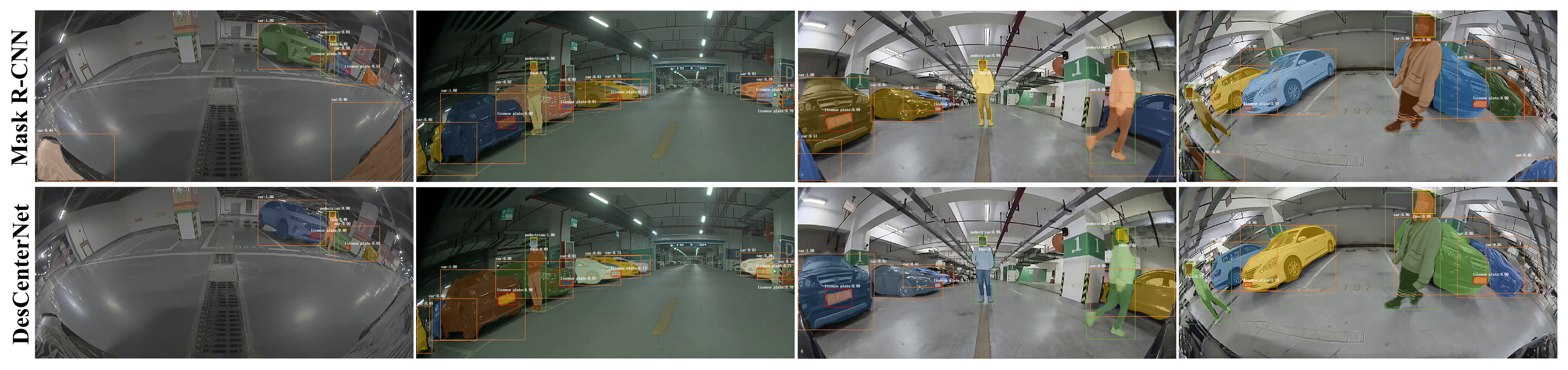}
\caption{Mask R-CNN (top) vs. DesCenterNet (bottom, Hourglass-104). Mask R-CNN has a false detection problem at the edge of the image.}
\label{fig:vsmaskrcnn}
\end{figure*}

\section{Methodology}
In this section, we introduce a new end-to-end face and vehicle license plate desensitization framework, which can be taken as a baseline model for future desensitization research.

\subsection{Overall Framework}
\begin{figure*}
\centering
\includegraphics[width=1.0\textwidth]{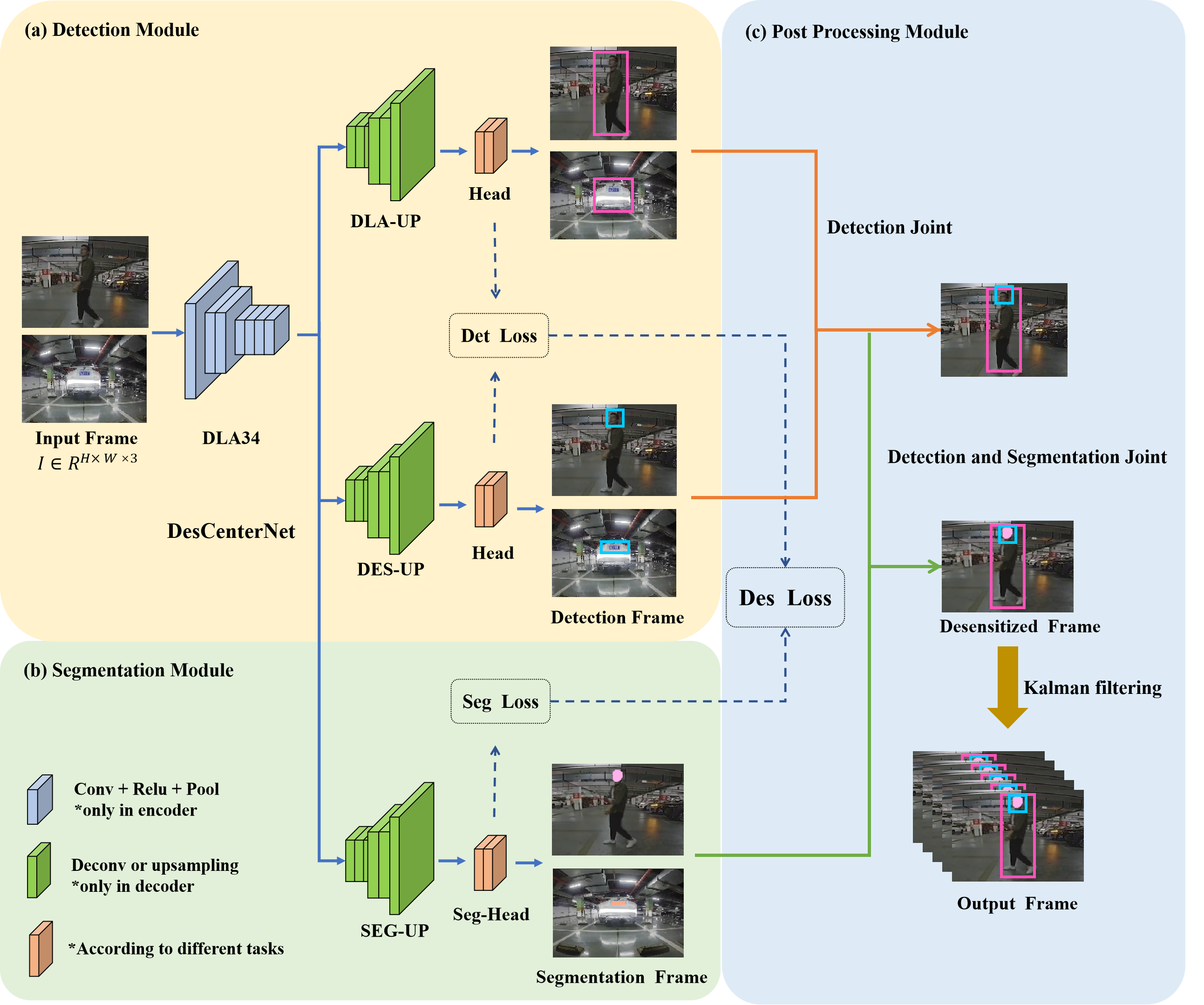}
\caption{Overview of our desensitization model. This is a two-stage model including (a) a detection module, (b) a segmentation module, and (c) a post-processing module. In the first stage, DesCenterNet consists of the backbone(DLA34), DLA-UP, and Head, which are the same as CenterNet \cite{bib68}. We add special DES-UP and Head for face and car-plate object detection. With multi heads, our DesCenterNet can predict the general object, vehicle license plate, and face region. On the other hand, for the location of the desensitization region, we will mark them by a small segmentation network, including SEG\_UP and Seg\_Head, and output the frame after desensitization. Finally, we provided a post-processing module that combined the joint desensitization method with Kalman filtering to improve face and vehicle license plate desensitization accuracy.}
\label{fig:overview}    
\end{figure*}

Fig. \ref{fig:overview} shows our framework design where
we desensitize faces, and vehicle license plates based on the multitask CenterNet \cite{bib68} model. The backbone network uses the DLA34 network \cite{bib68}. Different from the CenterNet model, we propose a new desensitization module (in Section 4.2) to achieve face and vehicle license plate desensitization tasks. 

First, we resize the input image to 640×480 and fully extract the object features through the DLA34 encoder, whose parameters are shared by the following three tasks. These tasks are comprised of face \& plate detection, face \& plate segmentation, and pedestrian \& vehicle detection, aiming to promote desensitization performance. In addition, to further address the problem of miss \& negative desensitization, we also add a post-processing module that will be discussed in detail in the following section.

\subsection{Desensitization Method}
Our desensitization model is a two-stage model.
In the first stage, our DesCenterNet serves for the detection of vehicle license plates or faces. The second stage is a small network FCN \cite{bib83} for segmentation of the desensitization region and then mosaicing them. 
After that, we can obtain the final desensitized output. 
The total pipeline of desensitization of the face is shown in Fig. \ref{fig:overview}.

\subsubsection{DesCenterNet}
\begin{figure}[t]

\includegraphics[width=0.5\textwidth]{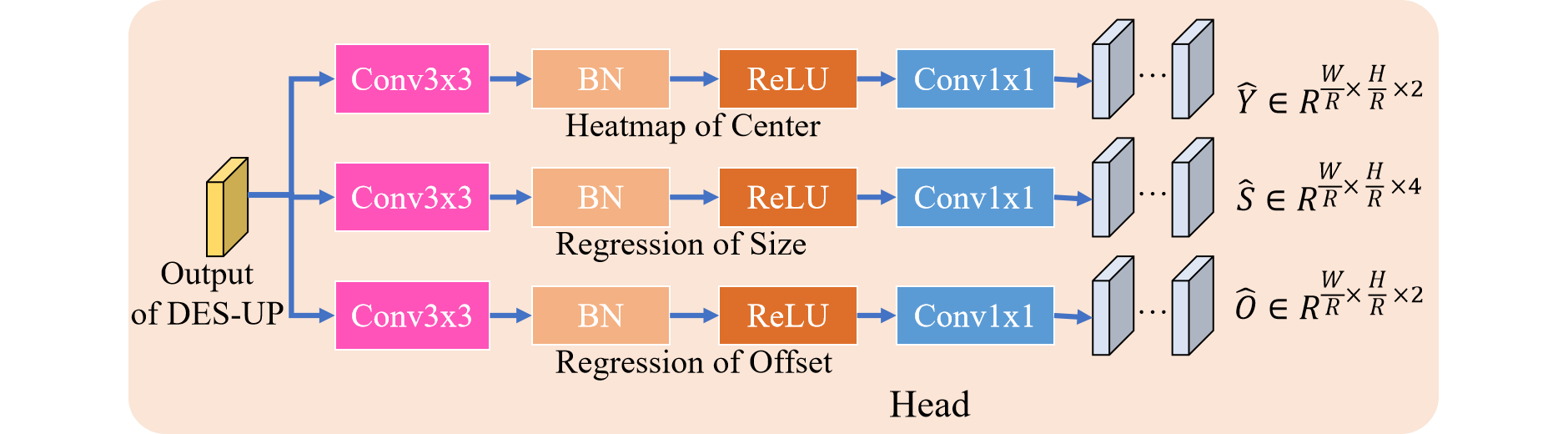}
\caption{Overview of our desensitization head. Inspired by CenterNet, we propose new heads for the detection of desensitization regions. Our heads consist of regression of a heatmap of the center, size of the object, and offset.}
\label{fig:overview_head}    
\end{figure}

In the first stage, DesCenterNet consists of the backbone(DLA34), DLA-UP, Head, Des-UP, and Head. The module DLA-UP and Head server predict cars and pedestrians, the same as CenterNet \cite{bib68}. Based on them, we add special DES-UP and Head for face and car-plate object detection. As shown in Fig. \ref{fig:overview_head}, our heads consist of regression of the heatmap of the center, size of the object, and offset. The desensitization task requires only the classification of the desensitized region and background. This task can be viewed as a two-classification task, and the output channel of the heatmap is set as 2. Furthermore, with multi heads, our DesCenterNet can predict general objects, vehicle license plates, and face regions.

\subsubsection{Segmentation Module}
The second stage is a segmentation model for the pixel classification of the desensitization region(face and vehicle license plate) and background. Its input is a feature map extracted through the DLA34 network, and the network weights are shared. Learning the feature information of facial and license plate regions is of great help in extracting semantic information from sensitive regions. We choose a simple and effective FCN model to find the desensitization region and mosaic it. As shown in Table \ref{tab:segmentation _result}, the FCN achieves the best trade-off of speed and precision. This conforms to the requirements of real-time autonomous driving.

\subsubsection{Post-processing Module}
This work aims to conceal the sensitive area as much as possible while minimizing error detection. Meanwhile, 100\% desensitized using only the detection and segmentation results can be very hard. As a result, we propose a novel post-processing module to ensure performance. The module consists of a local \& global detection sub-module, detection \& segmentation sub-module, and Kalman filtering sub-module that utilizes multitasks results from the network and outputs the final result jointly.

\begin{figure}[t]
\includegraphics[width=0.5\textwidth]{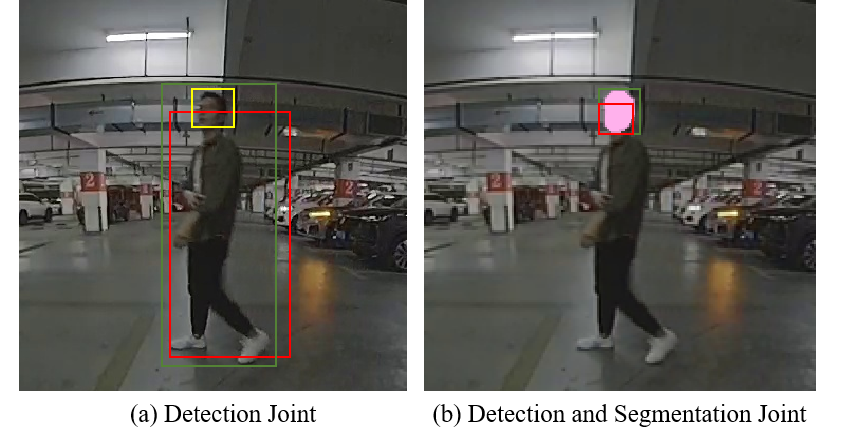}
\caption{Filtration operation of combined desensitization. \textcolor{green}{Green} represents an excellent detection box, red represents a poor detection box, \textcolor{yellow}{yellow} represents a face detection box, and \textcolor{pink}{pink} represents a face segmentation area. (a) Detection joint desensitization rule between pedestrian detection frame and face detection frame. (b) It is a joint desensitization rule between the detected joint desensitization and face segmentation.}
\label{fig:joint}    
\end{figure}

Regarding the local \& global detection sub-module, we aim to promote the performance by using both local \& global results (face \& person, plate \& vehicle). For instance, we first use both face and pedestrian detection results to overcome the face miss detection, Furthermore, to overcome the mistake detection, we output those face predictions with low confidence by checking if the face boxes are within the person box. The joint desensitization rules are shown in Fig. \ref{fig:joint}. It is worth noting that we treat the pedestrian detection box and the face detection box as a set of detection pairs and calculate their iou sequences. The red pedestrian detection box and the yellow face detection box in Fig. \ref{fig:joint} are considered unqualified detection pairs if their iou is less than 0.5, while the green pedestrian detection box and face detection box are considered qualified detection pairs and serve as the final output.

In addition, the detection \& segmentation sub-module further improves the performance. For the detection \& segmentation sub module, there are two specific methods. One is to obtain the minimum bounding box from the segmented area and perform iou calculation with the detection area, which can filter out some negative examples. Another approach that We output the final result jointly by considering the confidence of the same area on both tasks. We accept it when both are high and drop it when the results are relatively low. For example, when the detection confidence is greater than 0.7 and the segmentation iou is greater than 0.7. In this paper, we use the first method. Following this, a Kalman filtering algorithm is performed to smooth the result. 

\subsection{Desensitization Loss}
Inspired by the loss function of CenterNet, we proposed our loss function of desensitization, which consists of loss of keypoint heatmap, L1 loss of size, and offset. Our model will also output the keypoint heatmap $\hat{Y} \in [0,1]^{\frac{W}{R}\times \frac{H}{R} \times C}$, where the $R$ is the output stride, and $C$ is the number of keypoint classes. In the desensitization task, the $C$ is the 3, person's face, car plate, and background. According to the literature \cite{
bib91}, the down sampling of the output prediction factor R is 4. We select the focal loss \cite{bib91} in Equation \ref{eq:hm} to compute the distance between the predicted heatmap $\hat{Y}$ and the ground truth heatmap ${Y} \in [0,1]^{\frac{W}{R}\times \frac{H}{R} \times C}$, on which we transfer each target keypoint using a Gaussian kernel. The $\hat{Y}_{x,y,c}=1$ indicates the predicted keypoint, while the $\hat{Y}_{x,y,c}=0$ is the background. Thus, the training object can be viewed as a penalty-reduced pixel logistic regression with focal loss \cite{bib91}:
\begin{equation}
L_{k}=\frac{-1}{N} \sum_{x y c}\left\{\begin{array}{cl}
\left(1-\hat{Y}_{x y c}\right)^{\alpha} \log \left(\hat{Y}_{x y c}\right) & \text { if } Y_{x y c}=1 \\
\left(1-Y_{x y c}\right)^{\beta}\left(\hat{Y}_{x y c}\right)^{\alpha} & \\
\log \left(1-\hat{Y}_{x y c}\right) & \text { otherwise }
\end{array}\right.
\label{eq:hm}
\end{equation}

where $\alpha$ and $\beta$ are hyperparameters of the focal loss, and $N$ is the number of key points in the input image $I$. Following CenterNet \cite{bib68}, we use  $\alpha=2$ and $\beta=4$.

The output stride will cause the loss of the decimal part during the down-sampling. To recover this discretization error, we import the local offset loss \cite{bib91}:
\begin{equation}
L_{off}=\frac{1}{N} \sum_{p} \vert \hat{O}_{\tilde{p}}-\left(\frac{p}{R}- \tilde{p} \right) \vert
\end{equation}
, where $\hat{O}_{\tilde{p}}$ is predicted offset. Moreover, $p \in \mathcal{R}^2 $ of class $c$ is the ground truth point, and its location after down-sampling in output is $\tilde{p}=\lfloor \frac{p}{R} \rfloor$. The difference between them is the target offset.

After the prediction of the center point, we need the regression of the size of objects by the L1 loss:
\begin{equation}
L_{size}=\frac{1}{N} \sum_{k=1}^{N} \vert \hat{S}_{p_{k}}-s_{k} \vert. 
\end{equation}
 where $\hat{S}_{p_{k}} \in \mathcal{R}^{\frac{W}{R}\times \frac{H}{R} \times 2}$ is the predicted size and $s_k=(x^{k}_2 - x^{k}_1, y^{k}_2 - y^{k}_1)$ is the ground truth size of object $k$ and is created by target bounding box  $(x^{k}_1, y^{k}_1, x^{k}_2 - y^{k}_2)$.

To predict the region of desensitization, we import the loss of segmentation in equation \ref{eq:seg}. The ground truth mask is ${T}_{xyc} \in [0,1]^{\frac{W}{R}\times \frac{H}{R} \times 3}$, and its predicted semantic map is $\hat{T}_{xyc}$. Following the keypoint map loss, we also use focal loss \cite{bib91} to balance between hard and easy examples. We set $\alpha=2$ and $\beta=4$.
\begin{equation}
L_{seg}=\frac{-1}{N} \sum_{x y c}\left\{\begin{array}{cl}
\left(1-\hat{T}_{x y c}\right)^{\alpha} \log \left(\hat{T}_{x y c}\right) & \text { if } T_{x y c}=1 \\
\left(1-T_{x y c}\right)^{\beta}\left(\hat{T}_{x y c}\right)^{\alpha} & \\
\log \left(1-\hat{T}_{x y c}\right) & \text { otherwise }
\end{array}\right.
\label{eq:seg}
\end{equation}

After the computation of all losses of multi heads, our model requires the overall loss function:
 \begin{equation}
L= L_{k} + \lambda_{off} L_{off} + \lambda_{size}L_{size} + \lambda_{seg}L_{seg}.   
\end{equation}
where we set $\lambda_{off}=1$, $\lambda_{size}=0.1$ and  $\lambda_{seg}=0.5$ in all experiments and desensitization taska. Our network will predict the keypoint heatmap $\hat{Y}$, offset $\hat{O}$, size $\hat{S}$ and pixel-level class $\hat{T}$. They all share a common backbone network, such as DLA34 \cite{bib68}.

\section{Experiments}
In this section, we validate DesCenterNet's strength on our \textbf{ADD} dataset. We also perform extensive ablation and contrast experiments to demonstrate the effectiveness of the desensitization framework proposed in this paper.


\subsection{Experimental Setup}
The experimental procedure is set up under the Ubuntu 18.04.06 system, with python 3.8.5 and Pytorch 1.8. We train the \textbf{mixD} Dataset with 120 epochs. The learning rate is 0.12, momentum is 0.9, the learning decay rate is 0.001, the optimizer is SGD \cite{ruder2016overview}, and the batch size is set to 32. It is worth noting that this experiment uses the mixD dataset, which will be open-sourced after publication with another two sub-datasets.
%
\subsection{Main Results}
\begin{figure*}
\centering
\includegraphics[width=0.98\textwidth]{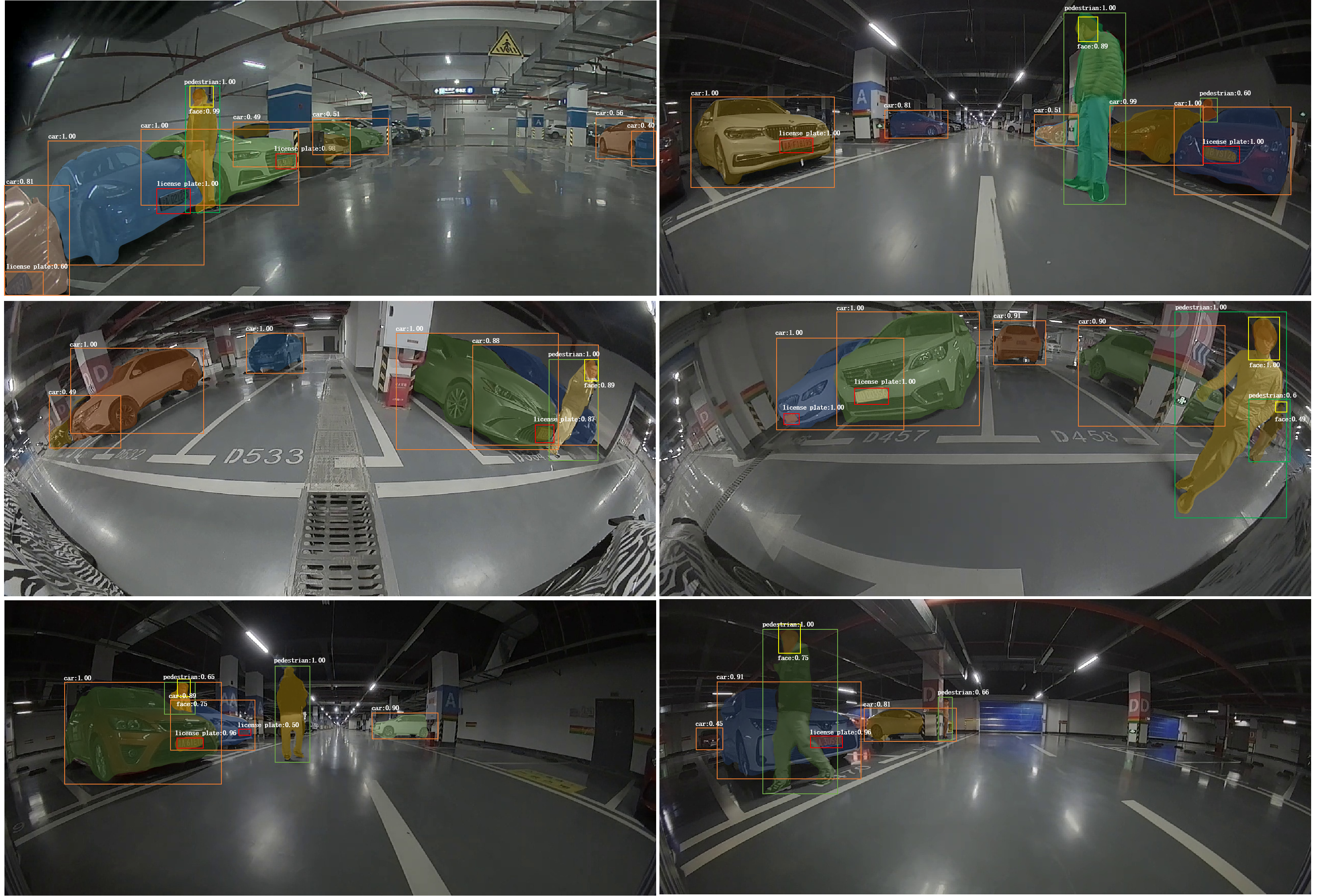}
\caption{Visualized results of our desensitization method on \textbf{ADD's mixD} test set. \textcolor{green}{Green} represents the pedestrian detection frame, \textcolor{orange}{orange} represents the car detection frame, \textcolor{red}{red} represents the license plate detection frame, and \textcolor{yellow}{yellow} represents face detection frame. Other masks, such as pedestrian, car, face, and vehicle license plate, are masks after segmentation.}
\label{fig:Visualization results}   
\end{figure*}

We compare DesCenterNet to the state-of-the-art methods in instance segmentation in Table \ref{table:vsmaskrcnn}.
All instantiations of our model outperform baseline variants of previous state-of-the-art models, including Mask R-CNN\cite{bib1037} and FCIS\cite{bib1038}. There are many improvements on Mask R-CNN, and we also hope there can be more improvements beyond the scope of work.
DesCenterNet outputs are visualized in Fig. \ref{fig:Visualization results}. DesCenterNet achieves good results even under challenging conditions. In Fig. \ref{fig:vsmaskrcnn}, we compare DesCenterNet and Mask R-CNN. Mask R-CNN is prone to false detection in the image edge. DesCenterNet shows no such situation. It is worth noting that all experiments maintain consistent training and evaluation protocols. When training Mask R-CNN, replace the segmentation head with Seg-UP and Seg-Head to obtain segmentation results of the same resolution size.

\begin{table*}[!t]
\centering
\caption{Our method DesCenterNet $vs$ Mask R-CNN with bounding boxes AP and IOFF.}
\label{table:vsmaskrcnn}
\begin{tabular*}{\tblwidth}{@{}LLLLLLLL@{}}
\hline
Method                & backbone        & $AP$           & $AP_{50}^{bb}$          & $AP_{75}^{bb}$         & $mIOFF$         & $IOFF_{50}$       & $IOFF_{75}$        \\ \hline
FCIS\cite{bib1038} +OHEM    & ResNet-101-C5-Dilated   & 39.1          & 59.8          & 39.8          & 42.2          & 56.2          & 45.7          \\
FCIS+++\cite{bib1038} +OHEM   & ResNet-101-C5-Dilated   & 42.2          & 62.7          & 44.9          & 48.1          & 62.1          & 51.2          \\
Mask R-CNN \cite{bib1037}            & ResNet-101-C4   & 42.1          & 64.8          & 45.9          & 49.2          & 61.2          & 52.7          \\
Mask R-CNN \cite{bib1037}            & ResNet-101-FPN  & 43.2          & 65.6          & 46.1          & 52.1          & 65.1          & 55.1          \\
Mask R-CNN \cite{bib1037}            & ResNeXt-101-FPN & 45.0          & 66.9          & 47.9          & 54.1          & 68.8          & 59.2          \\ \hline
\textbf{DesCenterNet} & ResNet-101      & 46.1          & 66.9          & 49.1          & 56.2          & 70.4          & 62.7          \\
\textbf{DesCenterNet} & Hourglass-104   & \textbf{48.7} & \textbf{70.1} & \textbf{49.2} & \textbf{61.0} & \textbf{70.8} & \textbf{65.7} \\ \hline
\end{tabular*}
\end{table*}

\subsection{Ablation Experiments}
\subsubsection{Comparison with different backbones} 

\begin{table}[h]
\small
\centering

\caption{Speed/accuracy trade-off for different backbones on the validation set.}
\label{tab:methods} 
\begin{tabular}{c|cccc}
    \toprule
	 Backbone & $mIOFF$ & $IOFF_{50}$ & $IOFF_{75}$ & $FPS$  \\
    \midrule
     Hourglass-104 &  61  & 70.8 & 65.7  & 14  \\
     DLA-34 &  55  & 72.9 & 60.3  & 52  \\
     ResNet-101 &  56.2  & 70.4 & 62.7  & 60  \\
     ResNet-18 &  51.2  & 60.4 & 55.7  & 153  \\
   \bottomrule
\end{tabular}
\end{table}
We use CenterNet\cite{bib68} as the baseline and propose Des-CenterNet to incorporate an MDN to achieve face and vehicle license plate desensitization tasks. Table \ref{tab:methods} shows the results of our validation set with different backbones. The FPS has been tested on our embedding machine, and the maximum computing capacity is approximately 1 Tera Operations Per Second (TOPS), much smaller than a standard Nvidia GPUS (32 TOPS).  Hourglass-104 achieves the best accuracy at a relatively good speed, with a 61\% mIOFF in 14 FPS. Using ResNet-101, we achieve the minimum real-time requirement for autonomous driving and outstanding performance with 56.2\% mIOFF.


\subsubsection{Comparison with different segmentation models}
Here, we study the results from different segmentation modules, including FCN \cite{bib83}, SegNet \cite{bib86}, PSPNet \cite{bib87}, RefineNet \cite{bib88}.
%
The metric of segmentation is the same as the PASCAL VOC 2012. 
%
This result is shown in Table \ref{tab:segmentation _result}, and RefineNet achieved the highest performance with $IOFF_{50}$ of up to 82.4\%. However, its speed is only 10 FPS, which is the lowest. In addition, FCN can infer 14 FPS under the same inputs, which is the fastest compared to others.
Because the desensitization task requires the inference model's high speed, we choose the simplest FCN as the segmentation model after trade-off performance and speed. 


\begin{table}[h]
\caption{The comparison of different segmentation models on our desensitization dataset.}
\label{tab:segmentation _result}
\centering

\begin{tabular}{c|c|cc}
\toprule
Dataset &Method & $IOFF_{50}$    & $FPS$    \\ \midrule
\multirow{7}{*}{mixD} &FCN-8s \cite{bib83}        & 70.8     & 14          \\
&SegNet \cite{bib86}            & 69.82     & 12         \\
&DeepLab  \cite{bib85}           & 70.6    & 7        \\
&PSPNet \cite{bib87}             & 67.4     & 6    \\
&RefineNet \cite{bib88}           & 82.4    & 10         \\ 
\bottomrule
\end{tabular}
\end{table}

\subsubsection{The importance of joint desensitization} 

\begin{table}[h]
\caption{Different combined desensitization methods of ablation experiments.}
\label{tab:importanceofjoint}
\centering
\begin{tabular}{ccccccc}
\hline
Dataset              & $DJ$ & $DSJ$ & $KFJ$  & $mIOFF$ & $IOFF_{50}$ & $IOFF_{75}$ \\ \hline
\multirow{6}{*}{mixD} & -                & -         & -                         & 60.9  & 70.8   & 65.3   \\
                     & \checkmark                & -      & -                            & 61.0  & 70.8   & 65.8   \\
                     & -                & \checkmark          & -                        & 65.9  & 71.5   & 66.3   \\
                     & -                & -          & \checkmark                        & 65.3  & 71.0   & 66.0   \\
                     & \checkmark                & -      & \checkmark                            & 66.0  & 71.4   & 66.4   \\
                     & -                & \checkmark     & \checkmark          & \textbf{69.2}  & \textbf{74.0}   & \textbf{70.1}   \\ \hline
\end{tabular}
\end{table}
To cover up the sensitive parts of the face and license plates as much as possible, we proposed a joint desensitization method in Section 4.2 and refined the combined desensitization method into three modules. One is a joint detection module (DJ) for pedestrian and vehicle detection boxes, faces, and license plates. The other is the joint of the result obtained by DJ and the segmentation module (DSJ). The last module is the Kalman filter joint module (KFJ). To verify the effectiveness of different joint methods, we conducted ablation experiments with different joint desensitization methods, as shown in Table \ref{tab:importanceofjoint}. In particular, we named only the DJ method w/ DJ, and the method combined with DJ and KFJ was named w/ DJ\&KFJ. Others were the same.

We use the method without the combined desensitization module as the baseline for ablation experiments. From Table \ref{tab:importanceofjoint}, we can see that the joint desensitization module's method has achieved good performance improvement. With only the DJ methods, mIOFF and $IOFF_{75}$ increased by 0.1\% and 0.5\%, respectively, but the improvement effect was small. Our analysis shows that the return accuracy of the detection box is significant, and the overlap rate of the detection box is significant, but the desensitization accuracy is limited. In the method of using DSJ, we can see that mIOFF, $IOFF_{50}$, and $IOFF_{75}$ have increased by 5.0\%, 0.7\%, and 1.0\% compared with baseline, respectively. The facial and license plate characteristics obtained by the segmentation module can effectively filter the noise in the predictive information, improving the recognition rate and desensitization accuracy of sensitive information. We used DSJ and KFJ in combination and found that $IOFF_{75}$ increased by 3.7\%, reaching the highest score. DSJ can effectively detect sensitive information, and KFJ can optimize the missed detection rate, significantly improving desensitization performance.


\begin{figure*}
\includegraphics[width=1.0\textwidth]{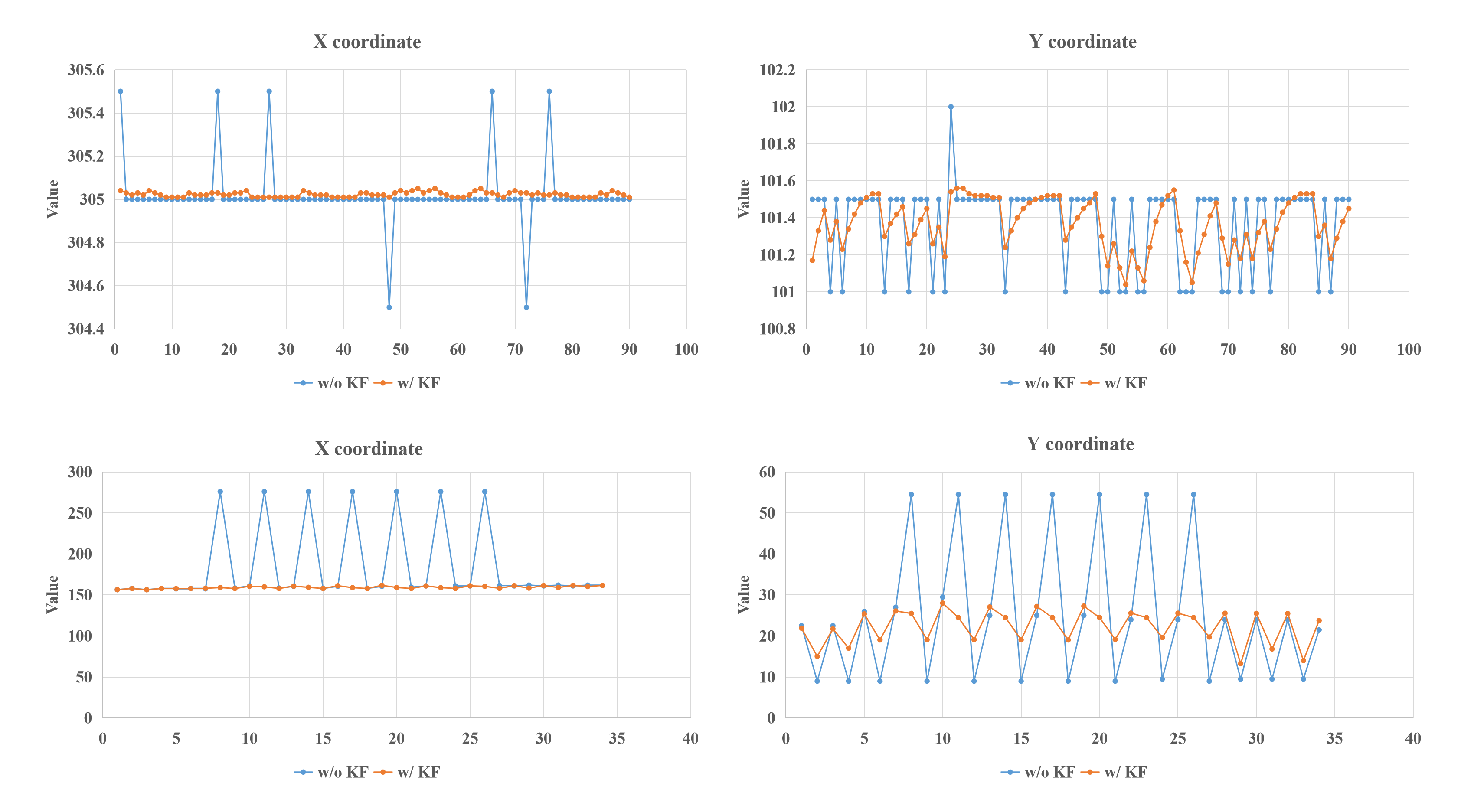}
\caption{The center point coordinate renderings of with Kalman Filter(w/ KF) and without Kalman Filter(w/o KF), the first row is the rendering of the center point of the vehicle license plate, and the second row is the rendering of the center point of the face.}
\label{fig:KF}    
\end{figure*}

\subsubsection{The importance of Kalman filter} 
\textbf{Indicator improvement.} As shown in Table \ref{tab:importanceofjoint}, in w/ KFJ method, mIOFF, $IOFF_{50}$, and $IOFF_{75}$ increased by 4.4\%, 0.2\%, and 0.7\% compared with Baseline, respectively. It can be seen that KFJ can effectively optimize the missed inspection effect.

However, compared to w/ DSJ method, mIOFF was reduced by 0.6\%, and mIOFF was increased by 4.3\%compared to the w/ DJ method. DJ can detect face and license plate information but also cover useless background information. DSJ can enhance the detection and desensitization effect of face and license plates from pixel-level and characteristic levels. KFJ is a relatively lightweight tracking method.

Finally, we combined KFJ, DJ, and DSJ respectively and found that w/ DJ\&KFJ and w/ DSJ\&KFJ increased by 0.7\% and 3.9\% compared with the mIOFF of w/ KFJ, respectively, to the highest. DSJ can effectively detect sensitive information, and KFJ can optimize the missed inspection rate. The two complement each other, which significantly improves desensitization performance.

\textbf{Stability and smoothness.} We highlight the critical performance of Kalman filtering in the post-processing in Fig. \ref{fig:KF}. Among the desensitization results obtained, the post-treatment of Kalman filtering makes the deviation of the center point of the face and the center of the license plate smaller, and the desensitization performance is more stable.

\subsection{Different Kalman filter previous frame selection}
To promote the smoothness of the desensitization task, we apply Kalman filtering proposed by \cite{bib90}, which aims to predict the target's location using previous frames when the network miss-detects the objects. In this section, we study how many previous frames should be included to predict the target because they determine the target motion, while too many frames cause false detection cases and insufficient frames caused missed detection cases. 

As shown in Table \ref{tab:kf}, we conducted ablation experiments with different previous frames and found that the best performance was achieved when the frame number was 4. We think that increasing the number of previous frames increases the model's robustness and the false detection rate. When the number of frames exceeds 4, the possible invalid frames will affect the desensitization effect, decreasing the IOFF index.
\begin{table}[!t]
\caption{Different Karman Filter previous frames selection.}
\label{tab:kf}
\centering
\begin{tabular}{ccccc}
\hline
Dataset              & Previous Frames & $mIOFF$ & $IOFF_{50}$ & $IOFF_{75}$ \\ \hline
\multirow{6}{*}{mixD} & 2               & 67.1  & 71.9   & 67.1   \\
                     & 3               & 68.9  & 73.5   & 68.3   \\
                     & 4               & \textbf{69.2}  & \textbf{74.0}   & \textbf{70.1}   \\
                     & 5               & 66.9  & 72.5   & 67.1   \\
                     & 6               & 65.1  & 70.9   & 65.2   \\
                     & 7               & 63.2  & 70.0   & 63.9   \\ \hline
\end{tabular}
\end{table}

\section{Discussion}
\subsection{Using desensitization data to train other tasks}
It is worth noting that training using desensitization data also has a positive effect on the performance of pedestrian and vehicle-related tasks since those targets are partially masked (face and license plate) and according to the copy-paste method proposed by \cite{bib1035} that these masked objects will promote the robustness of the model. Furthermore, to achieve more effective desensitization tasks, we deploy a multitask training procedure to ensure that the shared backbone can still extract the correct features of sensitive areas.

\subsection{Performance of desensitization}
According to the regulation \cite{bib1036}, the sensitive information is erased as long as masks cover more than 50\% of the area of the sensitive objects. This gives a fault tolerance to the algorithm.
The performance of our proposed desensitization method mainly depends on the effectiveness of detection and segmentation, as the subsequent joint desensitization methods refer to their results, which has a high demand for detection and segmentation networks.

\subsection{Compare with instance segmentation}
The desensitization method is similar to the instance segmentation method, but the difference lies in the desensitization region division and weight setting of the face in the label (as shown in Table 3). In addition, we also propose new evaluation indicators for desensitization tasks and the joint desensitization method mentioned in the post-processing module. In fact, our method will obtain the desensitized area and then cover it with cartoon icons or gray rectangles. We believe that desensitization representations can be freely chosen, and there will be more details to be defined and refined for future desensitization tasks.

\subsection{Compare with other datasets}
Autonomous driving data desensitization is a relatively new topic, which requires that autonomous driving vehicles can effectively process private data. ADD is the first fisheye dataset for the study of autonomous driving data desensitization. It has a strong purpose and is different from other open data sets. Of course, other datasets can also achieve data desensitization tasks theoretically by additional annotations and methods, but this can not deny our original intention and contribution to the research of autonomous driving data desensitization. We also hope that more researchers will pay attention to the research direction of autonomous driving data desensitization.

\section{Conclusion}

This paper presents the first autopilot desensitization dataset, dubbed \textbf{ADD}, which includes face and license plate information. By providing face and vehicle license plates with distinct characteristics, \textbf{ADD} intends to aid the industry in the development of a safer advanced autonomous driving system. In addition, we proposed a new framework for desensitization, and extensive experiments demonstrated the superiority and generality of our method. In the future, we hope that \textbf{ADD} will stimulate more correlation with desensitization research and promote the protection of more sensitive information for industry applications.

\bibliographystyle{plain}
\balance
\bibliography{cas-refs}

\end{document}